\documentclass[10pt,twocolumn,letterpaper]{article}

\usepackage{cvpr}              % To produce the CAMERA-READY version
\usepackage[accsupp]{axessibility} % Improves PDF readability for those with disabilities.
\usepackage[dvipsnames]{xcolor}

\definecolor{cvprblue}{rgb}{0.21,0.49,0.74}
\usepackage[pagebackref,breaklinks,colorlinks,citecolor=cvprblue]{hyperref}

\def\paperID{52} % *** Enter the Paper ID here
\def\confName{CVPR}
\def\confYear{2024}

\title{IrrNet: Advancing Irrigation Mapping with Incremental Patch Size Training on Remote Sensing Imagery}

\author{Oishee Bintey Hoque\(^{1,2}\), Samarth Swarup\(^2\), Abhijin Adiga\(^2\), Sayjro Kossi Nouwakpo\(^3\), Madhav Marathe\(^{1,2}\)\\
\(^1\)Department of Computer Science, University of Virginia, Charlottesville, VA, USA\\
\(^2\)Biocomplexity Institute, University of Virginia, Charlottesville, VA, USA\\
\(^3\)US Department of Agriculture, Agricultural Research Service, Kimberly, ID, USA\\
{\tt\small \{oishee,swarup,abhijin,marathe\}@virginia.edu}, {\tt\small kossi.Nouwakpo@usda.gov}
}

\begin{document}
\maketitle
\begin{abstract}
Irrigation mapping plays a crucial role in effective water management, essential for preserving both water quality and quantity, and is key to mitigating the global issue of water scarcity. The complexity of agricultural fields, adorned with diverse irrigation practices, especially when multiple systems coexist in close quarters, poses a unique challenge. This complexity is further compounded by the nature of Landsat's remote sensing data, where each pixel is rich with densely packed information, complicating the task of accurate irrigation mapping. In this study, we introduce an innovative approach that employs a progressive training method, which strategically increases patch sizes throughout the training process, utilizing datasets from Landsat 5 and 7, labeled with the WRLU dataset for precise labeling. This initial focus allows the model to capture detailed features, progressively shifting to broader, more general features as the patch size enlarges. Remarkably, our method enhances the performance of existing state-of-the-art models by approximately 20\%. Furthermore, our analysis delves into the significance of incorporating various spectral bands into the model, assessing their impact on performance. The findings reveal that additional bands are instrumental in enabling the model to discern finer details more effectively. This work sets a new standard for leveraging remote sensing imagery in irrigation mapping.
\end{abstract}

\section{Introduction}
\label{sec:intro}
Irrigation mapping refers to the process of creating detailed maps to outline how irrigation systems (e.g., sprinkler, flood, drip irrigation etc.) are laid out across agricultural lands. Sustaining or enhancing irrigation is crucial for successful crop cultivation~\cite{ABIOYE2020105441} amid climate change-induced warmer temperatures, water scarcity~\cite{LEVIDOW201484} and shifting precipitation patterns~\cite{msu2020irrigation, epaNov2023}. However, the lack of comprehensive data detailing the timing, locations, and specific types of irrigation utilized poses a significant challenge for effective planning and management \cite{boutsioukis2022present}.

\begin{figure}
    \centering
    \includegraphics[width=.5\textwidth]{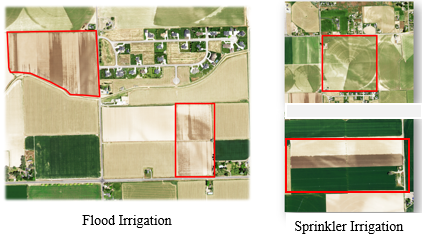}
    \caption{Representing the heterogeneity of irrigation types involves close proximity, forming diverse structures in the field characterized by variations in geometric shape, color, and patterns, despite exhibiting similarities.}
    \label{fig:irr_type}
\end{figure}

The availability of satellite images from platforms like Landsat and Sentinel, offering multiple spectral bands, including visible, near-infrared, and thermal infrared,has significantly facilitated the monitoring of agricultural landscapes on a large scale. Comparing to other dataset, Landsat holds the distinction of being the longest-running collection of moderate-resolution land remote sensing data obtained from space, providing continuous coverage over time\cite{usgs_landsat}. Geospatial semantic segmentation seeks to analyze these remote sensing imagery by assigning semantic labels to each pixel, enabling detailed interpretation of land cover and features. However, 
despite the abundance of raw satellite image data available today, for tasks like irrigation mapping remains challenging due to their scarcity of labels. Moreover, irrigation systems can vary widely in scale, from small-scale subsistence farming to large commercial agriculture. The heterogeneity in irrigation practices and systems across different regions adds to the complexity of mapping (see Fig. \ref{fig:irr_type}). In addition, the coarse resolution of Landsat imagery hampers models in capturing fine details, while in general remote sensing datasets exhibit substantial intra-class variance, adding to segmentation challenges\cite{app11115069,9857009}.

Recent advancements in deep learning-based segmentation models have greatly improved performance in segmenting agricultural features from high-resolution remote sensing images\cite{9857009,Marsocci_2023_CVPR,Bjorck2021AcceleratingES,10187150}. Transfer learning using semantic segmentation architectures such as U-Net\cite{unet}, DeepLabv3+\cite{deeplab}, FPN\cite{fpn}, SegFormer\cite{SegFormer} have shown promising results on remote sensing datasets\cite{Li_Wang_Zhao_Wang_Zhong_2023,10039322,10146268,ZHANG2023107511}. However, it is important to consider that the characteristics of agricultural objects, such as crops, fields, and irrigation systems, may differ from those of objects in traditional benchmark datasets. 1 pixel of Landsat image holds 30m area of information. In dense and cluttered agricultural landscapes, existing model encounters challenges in accurately mapping fields of various patterns and sizes. Standard segmentation models also struggle to effectively segment small-areas due to inadequate pixel representation, even in supervised conditions\cite{10208987}. Therefore, adapting and fine-tuning segmentation models to effectively capture these agricultural features is important for accurate mapping and analysis in agricultural applications.

Patch sizes are crucial parameters in remote sensing segmentation tasks. Studies shows that in the initial training phases, there's a notable correlation between gradients computed from small input patches and those from the entire image, fading as training progresses\cite{hoffer2019mix,NEURIPS2019_d03a857a}. Transformer based model leverages smaller patches from images to get better context,  with smaller patches leading to
higher accuracy\cite{dosovitskiy2021image}.
Many recent research have focused extensively on exploring the impact of varying patch sizes on segmentation accuracy \cite{10378441, Zhang2023,bioengineering10050534,Shen2023,TODESCATO2024121116,10205121}.

\begin{figure}
    \centering
\includegraphics[width=.45\textwidth]{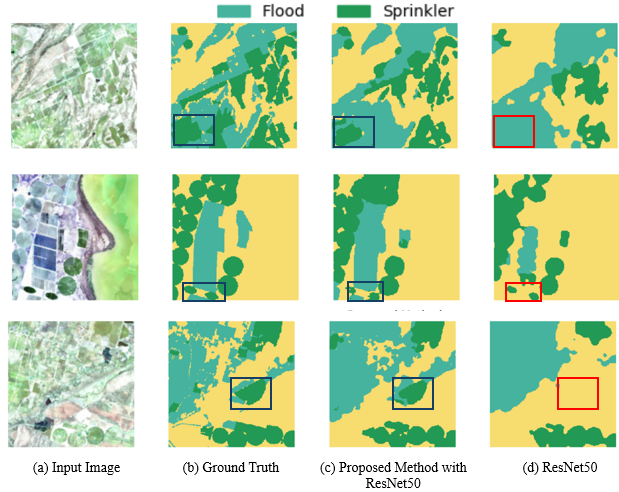}
    \caption{ Irrigation type predictions by a U-Net with ResNet50 as backbone and by integrating our proposed method. The predictions made by U-Net struggles to focus on fine details where by integrating proposed method the model performs better. Few areas have been marked in the image to show the better prediction by the models (red bounding boxes represent wrong prediction).
}
    \label{fig:exampleImprove}
\end{figure}

To address the issue of existing models failing to capture finer details due to inadequate pixel representation in larger patch settings, we propose a simple yet effective dynamic patch size utilization model in this work. This model employs encoder-decoder based architectures and dynamically adjusts patch sizes during training starting with smaller patches and gradually increases to larger patches, while sharing weights across different training stages. This approach aims to optimize performance and adapt to diverse landscape features effectively. By adjusting patch sizes at each stage, the model can focus on different levels of detail within the images, effectively leveraging limited labeled data to train large and data-intensive encoder-decoder models. This adaptability enables the model to better handle complex scenes with varying levels of detail, ultimately resulting in improved segmentation accuracy and generalization capability. Our proposed method has achieved around 20\% performance increase. The effectiveness of the proposed approach is demonstrated in Fig. \ref{fig:exampleImprove}, where the irrigation type predictions outperform those of previous methods.
In addition, we conducted further analysis on different channels to assess their impact on model learning. We also explored the integration of pretrained weights specifically trained on Landsat data to evaluate their effectiveness in enhancing performance. Furthermore, we compared our approach with existing state-of-the-art technologies to demonstrate its effectiveness. In summary, our contributions in this project encompass:

\begin{itemize}
    \item Proposed a novel model capable of mapping diverse irrigation methods within heterogeneous fields by leveraging varying patch sizes. To the best of our knowledge, no prior work has addressed irrigation type detection on such a large scale
    \item Demonstrated enhanced outcomes compared to the state-of-the-art, highlighting the influence of progressively training the models
    \item Identification of the impact and importance of various channels and patch sizes from satellite imagery to get a better performance
    \item Comparative analysis of the state-of-the art models to identify the most robust and effective solution

\end{itemize}

% In this project, our goal is to create a robust model capable of accurately identifying various irrigation methods used concurrently within a region. We plan to train the model to distinguish between irrigated and non-irrigated areas, with a particular emphasis on the characteristics of irrigated lands. This precision is vital for successfully identifying subtle features, especially in small sections where an irrigation type may exist within a predominantly different class or where rain-fed and irrigated agriculture co-exist, complicating the distinction between irrigated and non-irrigated fields\cite{DEINES2019111400}. To the best of our knowledge, this specific challenge has not been addressed in previous research.

% To achieve this we propose a self guided Multi-task learning segmentation model for irrigation mapping from satellite image. Utilizing a limited dataset from the Utah region, we will conduct an in-depth analysis of our model's performance and interpretability. Our primary goal is to build a model that achieves highly accurate detection of various irrigation types, while also uncovering the key features contributing to the differentiation of irrigation areas.  Additionally, we are committed to enhancing the model's generalizability to cover regions other regions different from Utah.

\section{Background}
Current irrigation mapping products provide spatial data without detailing irrigation methods \cite{salmon2015global, siebert2015globaldata}. While remote sensing has been utilized to map irrigated fields, especially in areas of mixed agriculture \cite{bazzi2019mapping}, distinguishing between irrigation types remains challenging due to landscape complexity and subtle practice variations. Computer vision and machine learning have made strides in identifying specific systems like center pivots \cite{geus2020fast,saraiva2020automatic, tang2021mapping}, or covering small area \cite{RAEI2022106977} but the broader task of irrigation mapping demands nuanced analysis.

Semantic segmentation focuses on predicting pixel-wise categories for input images. Over recent years, numerous approaches have been devised to facilitate real-time semantic segmentation. E-Net \cite{paszke2016enet} proposes a lightweight architecture tailored for high-speed segmentation. SegNet \cite{badrinarayanan2017segnet} combines a compact network architecture with skip connections to achieve rapid segmentation. ICNet \cite{zhao2018icnet} employs an image cascade algorithm to expedite the pipeline, while ESPNet \cite{mehta2018espnet,mehta2019espnetv2} introduces an efficient spatial pyramid dilated convolution. Furthermore, BiSeNet \cite{yu2021bisenetv2} segregates spatial details and categorical semantics to achieve a balance between high accuracy and efficiency in semantic segmentation. More recently, SegFormer \cite{xie2021segformer} incorporates Transformers with a multi-layer perceptron decoder, resulting in swift semantic segmentation. DeepLab V3+\cite{deeplab} enhances the DeepLab series through an encoder-decoder architecture, while newer approaches like OCR with HRNetV2\cite{wang2020deep} and SegFormer\cite{xie2021segformer} highlight the potency of Transformers in Semantic Segmentation tasks.
In contrast to conventional methods that predict distribution maps across classes for semantic masks, YOSO \cite{Hu_2023_CVPR} predicts kernels along with their corresponding categories for segmentation.

The role of image size in the training and evaluation of models is pivotal\cite{hoffer2019mix,VANNOORD2017583}. Techniques such as resizing and rescaling images are commonly employed for augmentation \cite{Shorten2019} or during the model training process\cite{TODESCATO2024121116}, with studies indicating that variability in image sizes can enhance model learning\cite{VANNOORD2017583}. The concept of "progressive resizing," as explored by \cite{karras2018progressive} and \cite{Howard2018FastAI}, involves incrementally increasing the size of images during training to bolster model performance and expedite convergence. Building on this, \cite{Touvron2019TrainTestResolution} showcased that Convolutional Neural Networks (CNNs) could be initially trained on a fixed, smaller image size and subsequently fine-tuned to a larger size for evaluation. This approach mitigates the discrepancy in performance due to changes in image size, facilitating quicker training times and heightened accuracy. In our study, we employ a progressive patch size increment technique, combined with encoder-decoder architectures and transfer learning strategies, to enhance irrigation mapping using remote sensing data.

\section{Methodology}
\begin{figure}
    \centering
\includegraphics[width=.5\textwidth]{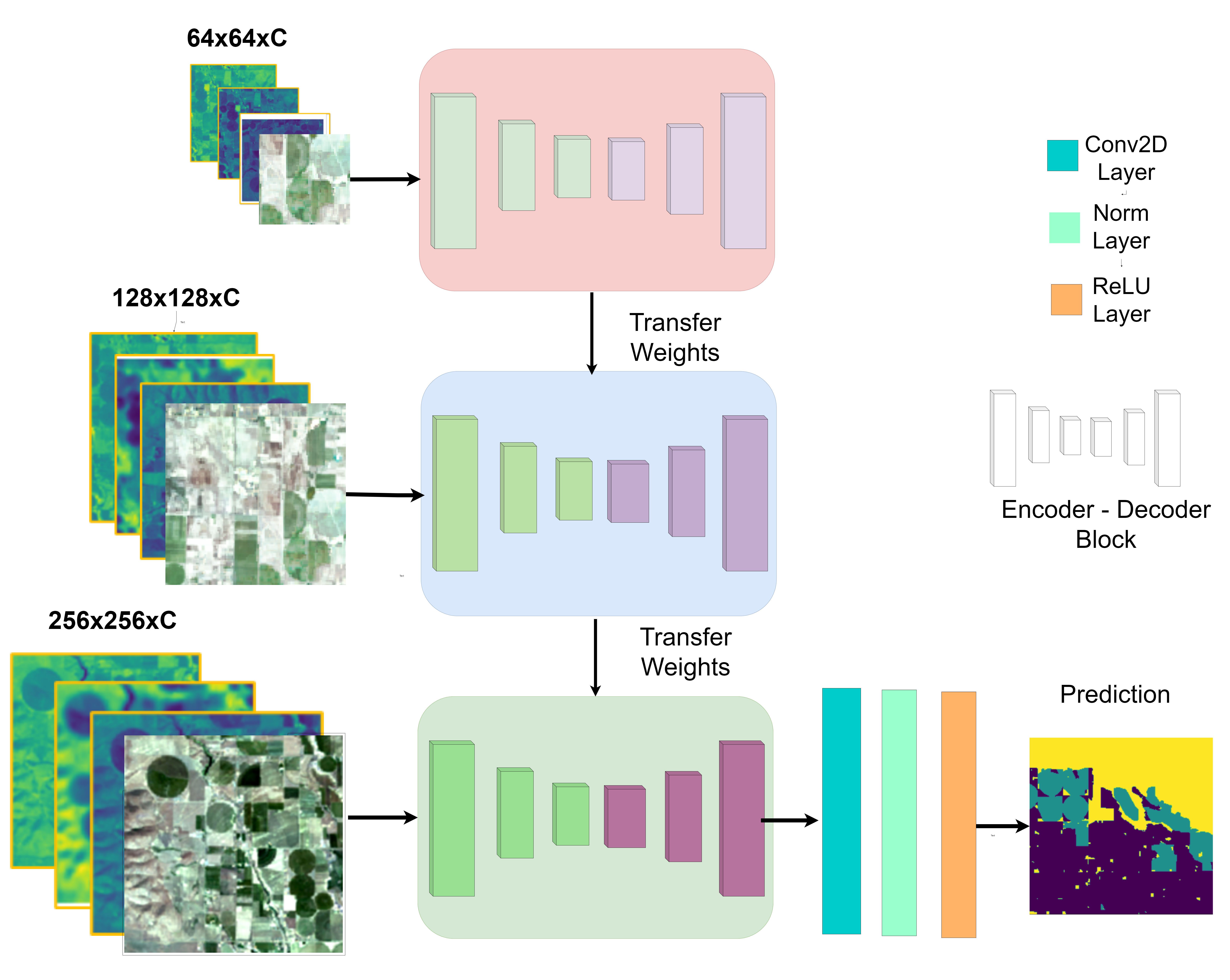}
    \caption{Illustration of the method proposed to integrate with the encoder-decoder based approach.}
    \label{fig:proposedMethod}
\end{figure}

Our methodology employs a sequence of innovative steps, combining progressive patch size training, multispectral channel analysis, encoder-decoder architectures with transfer learning, and a hybrid loss function approach.
Figure \ref{fig:proposedMethod} illustrates our overall approach, which we detail in
the following.

\subsection{Progressive Patch Size Training with Transfer Learning}

Our study introduces a targeted approach to map irrigation types, specifically distinguishing between sprinkler and flood irrigation systems through remote sensing imagery. The foundation of our method involves an initial training phase on small image patches ($P_{s}$) of dimensions $64 \times 64$ pixels. These patches are strategically selected to densely represent the class labels associated with the irrigation types, ensuring that the model ($M$) is exposed to pivotal data early in the training process
Initially, the model $M$ is trained on patches of size $P_{s}$ (e.g., $64 \times 64$ pixels), focusing on detailed feature extraction:

\[ M(P_{s}; \Theta_{s}) \rightarrow \Theta_{s+1} \]

where $\Theta_{s}$ and $\Theta_{s+1}$ represent the model parameters before and after training on $P_{s}$, respectively. Subsequently, the training progresses to larger patch sizes $P_{l}$ (e.g., $128 \times 128$, $256 \times 256$ pixels), utilizing the weights $\Theta_{s+1}$ as the initial parameters for training on larger patches. By incrementally increasing the patch size used in training phases, the model adapts to recognize patterns at varying scales, enhancing its predictive accuracy and reducing computational overhead.

\subsection{Transfer Weights on Spectral Bands}
In our approach to enhance model performance through additional spectral channels, we employed a transfer learning strategy that involved extending the model's existing weights to accommodate new channels. Let the original model trained on standard RGB channels be denoted as $M_{\text{RGB}}$, with its weights represented by $\Theta_{\text{RGB}}$. To incorporate new spectral bands, such as near-infrared (NIR) and short-wave infrared (SWIR), we expanded the input dimensionality, resulting in a modified model, $M_{\text{Extended}}$.

This process required adjusting the model's first convolutional layer to accept an increased number of input channels, $C_{\text{new}}$, where $C_{\text{new}} > 3$ due to the inclusion of channels beyond the traditional RGB. The augmentation of $\Theta_{\text{RGB}}$ to form a new set of weights, $\Theta_{\text{Extended}}$, integrates the learned features from the RGB model with the new spectral information. Mathematically, this adaptation can be described as:

\begin{equation}
    \Theta_{\text{Extended}} = \text{extend}(\Theta_{\text{RGB}}, C_{\text{new}})
\end{equation}

where the $\text{extend}(\cdot)$ function systematically enlarges the weight matrix to align with the augmented channel input. $M_{\text{Extended}}$ is subsequently fine-tuned on the dataset comprising the expanded spectral data, optimizing $\Theta_{\text{Extended}}$ to capitalize on the comprehensive spectral insights offered by the additional bands. This fine-tuning not only retains the knowledge from the original RGB channels but also exploits the unique characteristics of the new channels, significantly improving the model’s accuracy and robustness in irrigation pattern identification and classification.

\subsection{Hybrid Loss Function}
The model optimization employs a hybrid loss function, combining Binary Cross-Entropy (BCE) and Dice loss $L_{\text{Dice}}$ to effectively balance pixel-wise accuracy and region overlap:
\[ L = \alpha L_{\text{BCE}}+ \beta L_{\text{Dice}}\]
where $\alpha$ and $\beta$ are the weighting coefficients balancing the two loss components.

\section{Experiments and Results}
\subsection{Study Area}
The study utilizes the Utah Water Related Land Use (WRLU) dataset to develop and train a deep learning model for irrigation mapping. Utah covers an area of $219,807$ $km^{2}$ and spans longitudes from $109^\circ$W to $114^\circ$W and latitudes from $37^\circ$N to $42^\circ$N. Farming occupies $43,301$ $km^{2}$ of land, with $18\%$ of agricultural land being irrigated in Utah. Irrigation is more prevalent on moderately sized farms, typically ranging from $20$ to $22$ hectares. We utilized WRLU dataset covering the years 2003 to 2021, excluding 2016 due to the absence of digitized data. The study focused on agricultural areas within the dataset, categorizing irrigation methods into flood, sprinkler, drip, dry crop, and sub-irrigated, with sprinkler systems further divided based on their mobility. This research highlighted the prevalence of wheel line and center pivot sprinkler systems, along with surface irrigation techniques such as furrow and basin irrigation.

To evaluate our method in different area than the training, we use the data from southern Idaho's Upper Snake Rock (USR) watershed, overseen by the Twin Falls Canal Company (TFCC), spanning 6,300 $km^{2}$. With 37\% of land dedicated to irrigated agriculture mirroring Utah's irrigation methods, with nearly $80\%$ of larger farms receiving irrigation water. Additional labeling data was collected within and around this irrigation project area to supplement the WRLU dataset. An expert familiar with irrigation methods in the region labeled agricultural fields within 235 square polygons of size $1.9$ km. Irrigation methods were identified by examining field appearances and spatial patterns on high-resolution aerial imagery from the National Agriculture Imagery Program (NAIP). 

\subsection{Dataset Collection and Preparation}
\begin{figure*}
    \centering
\includegraphics[width=\textwidth]{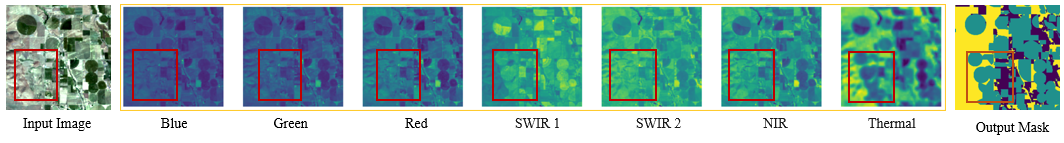}
    \caption{Visualization of each band (Blue, Green, Red, Shortwave Infrared (SWIR) 1, SWIR 2, Near Infrared (NIR), Thermal in an input image. An area is highlighted with a bounding box to exemplify how each band contains distinct information within an image.}
    \label{fig:7Bands}
\end{figure*}

The irrigation methods mapping tool utilizes Landsat Tier 1 Surface Reflectance data from Landsat missions 5 and 8 to construct input imagery. Landsat 5 covers the period from 2003 to 2011 and Landsat 8 from 2013 to 2021 with a repeat cycle of 16 days. The model inputs are computed over 4-month period, averaging each band,spanning from April 1st to September 1st. This corresponds to the period when irrigation is the most active in the region. The Google Earth Engine platform was utilized for acquiring Landsat imagery and processing model inputs. From 2003 to 2017, the annual WRLU survey extent encompassed only a portion of the total agricultural area footprint of the State of Utah, whereas WRLU maps after 2018 covered the entire state. Using the WRLU dataset, the irrigation methods in each patch are labeled as Flood (F), Sprinkler (S), or Other (O). Class O includes non-irrigated areas and those irrigated using less common methods like drip irrigation. To maintain class balance within the model, images containing more than 90\% other class pixels were omitted. These input images are standardized and clipped to 256x256x7 patch squares, each patch covering around 59$km^2$ of the study area at a 30 m resolution (see fig \ref{fig:7Bands}). Training and validation images are randomly selected, with training squares spatially distinct from validation squares.  In total, 798 training and 127 validation images were extracted for model development and evaluation.

\subsection{Data Preprocessing}
\begin{figure}
    \centering
\includegraphics[width=.45\textwidth]{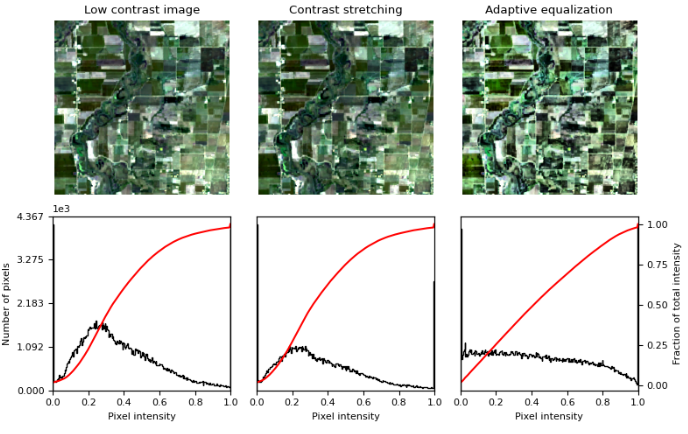}
    \caption{After performing CLAHE histogram equalization on the normalized image.}
    \label{fig:eq}
\end{figure}

Satellite images exhibit variability due to factors like lighting conditions and sensor characteristics. Normalization ensures consistency across different scenes and sensors by standardizing brightness and contrast. Initially, lower and upper percentiles are calculated to define the range of pixel values. This range is then used to clip and normalize the image, confining pixel values within the calculated bounds and scaling them to a range of 0 to 1. Remote sensing imagery can have varying atmospheric conditions, lighting, and land cover can obscure important details To address this issue, additionally, CLAHE histogram equalization\cite{6968381} is performed to enhance contrast (see fig \ref{fig:eq}).  One of the advantages of CLAHE over traditional histogram equalization is its ability to limit noise amplification. By applying a contrast limit to the histogram equalization process, CLAHE avoids the over-enhancement of contrast in near-uniform regions, which can often lead to noise becoming more pronounced, a common issue in remote sensing images due to sensor irregularities or atmospheric interference. Following this, we generate sets of 64x64 and 128x128 image patches extracted from 256x256 patches, while excluding patches where over $80\%$ of the pixels belong to other regions.

\subsection{Training and Experiments}
\label{sec:training}
This section provides a comprehensive overview of our experimental approach for developing an irrigation mapping model, detailing the methodology and evaluating the outcomes across three primary dimensions. We aim to illustrate the impact of utilizing various state-of-the-art (SOTA) models, the integration of multiple channels, and the application of progressive training techniques across different patch sizes at each stage of the model's training.\\
\textbf{Experimental Settings.} In summary, we leverage a suite of transfer learning with existing deep learning architectures as the foundational models for our experiments, specifically ResNet50, DeepLabv3+, SegFormer, InceptionV3 and pre-trained ResNet50 on Landsat dataset\cite{Stewart_TorchGeo_Deep_Learning_2022}. The original final layers of these models are replaced with a custom sequence of layers designed to enhance feature extraction and adaptability to our Landsat dataset. This sequence includes Conv2D with stride 1 as upsampling layers and three successive blocks composed of Conv2d, Batch Normalization (BN), Dropout, and ReLU layers. The configuration of these blocks varies: the first block adapts its size of inputs based on the chosen base model's architecture, while the last block outputs the segmentation mask.  Additionally, batch normalization is applied after the last block to enhance model performance by normalizing the final layer's output.

Our training process is divided into two distinct phases to optimize the integration of the base models with our custom layers. Initially, the base model is frozen, and only the appended layers are trained for two epochs. This approach allows the model to adjust the pre-learned features for the landsat dataset predictions. Following this, the entire model, including the previously frozen base layers, undergoes fine-tuning with variable learning rates to refine feature extraction across the model's depth. The learning rates are adjusted such that the earlier layers, which capture fundamental shape and feature information from the ImageNet dataset, receive a lower rate compared to the more specialized deeper layers. This differential learning rate strategy is set between 3e-3 and 3e-4 using a minibatch size of 16.

\textbf{Training Time Augmentation.}
To enhance the training process and ensure our model is robust to various image transformations, we incorporated several train-time augmentation techniques. These included flipping images horizontally and vertically to simulate different orientations, and rotating images within a range of -20\% to +30\% to account for variations in perspective. Additionally, we experimented with zooming in or out on the images by 20\% to 30\% and adjusting the brightness by up to 20\% to mimic different lighting conditions. Furthermore, we adjusted the contrast of the images by 20\% to simulate variations in image quality and applied translations up to 5\% in both the horizontal and vertical directions to represent slight shifts in the camera angle or object position. These augmentation strategies were carefully chosen to expose the model to a wide range of plausible variations it might encounter in real\-world scenarios, thereby improving its generalization capabilities and performance on unseen data.

\textbf{Evaluation Metrics.} In evaluating our model's efficacy for classification tasks, we utilized metrics including Mean Intersection over Union (mIoU), Precision, Recall, and F1-Score. These metrics collectively provide a multifaceted view of the model's performance, with mIoU assessing spatial accuracy through predicted and actual segment overlaps, and the other metrics gauging overall prediction accuracy, positive prediction correctness, capability to identify all relevant instances, and a harmonized measure of precision and sensitivity, respectively.

\[ \text{mIoU} = \frac{1}{N} \sum_{i=1}^{N} \frac{\text{TP}_i}{\text{TP}_i + \text{FP}_i + \text{FN}_i} \]

\[ \text{Precision} = \frac{\text{TP}}{\text{TP} + \text{FP}} \]
\[ \text{Recall} = \frac{\text{TP}}{\text{TP} + \text{FN}} \]
\[ \text{F1-Score (F1)} = 2 \times \frac{\text{Precision} \times \text{Recall}}{\text{Precision} + \text{Recall}} \]

\begin{figure*}
    \centering
\includegraphics[width=\textwidth]{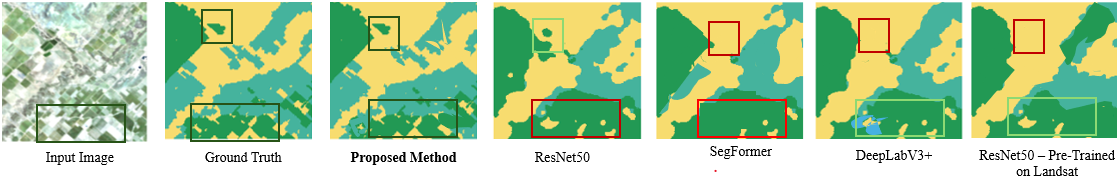}
    \caption{Results obtained from various state-of-the-art methods are compared with ours. Dark green bounding boxes highlight areas where our methods have outperformed the others. In this setup, the proposed method is trained using six channels, while the remainder of the model is trained using RGB.}
    \label{fig:allMethodResult}
\end{figure*}

\begin{figure}
    \centering
\includegraphics[width=.5\textwidth]{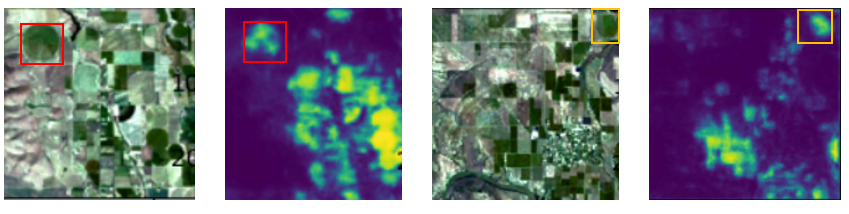}
    \caption{The above heatmaps belonging to sprinkler irrigation, we observed the limitation in the existing model's ability to comprehend patterns effectively. For instance, in the provided images, the model appears to prioritize color over shape, indicating a gap in its understanding of the underlying structures.}
    \label{fig:error}
\end{figure}

\begin{figure}
    \centering
    \includegraphics{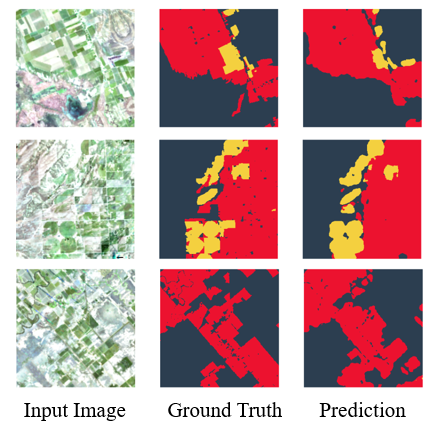}
    \caption{Visualization of Model's Prediction.(Red belongs to flood and yellow belongs to sprinkler irrigation.)}
    \label{fig:predOutput}
\end{figure}

\section{Ablation Study and Analyses}
This section is structured as follows: First, we examine the performance of SOTA methods within our specific setup, identifying areas where they excelled in generating irrigation maps and noting any challenges encountered. Next, we explore the potential for enhancing model performance through the integration of various spectral channels. Finally, we assess the impact of employing different patch sizes during training on the model's effectiveness.

\textbf{Performance of SOTA models.} Table \ref{table:SOTA} presents a detailed evaluation of SOTA models. In addition to employing pre-trained models from the ImageNet database, we also trained our dataset with a ResNet50 model pre-trained on Landsat 7 \& 8 data, utilizing 7 channels similar to our study. Our findings indicate that models pre-trained on Landsat data outperform others, which is expected since the ImageNet dataset, despite its size, differs significantly from Landsat in content. Furthermore, given that Landsat imagery has a 30m resolution, each pixel encompasses a considerable amount of information. A 256x256 image covers a vast area, containing an abundance of details. Directly training models with these larger images leads to difficulties in capturing the general patterns of different irrigation types, as illustrated in fig \ref{fig:error}.

\textbf{Impact of Different Bands.} Our initial experiments were conducted using the RGB bands. Irrigation mapping can rely on multiple factors including the color of the fields, patterns within the fields, and the level of greenness, among others, which may not be fully captured through RGB channels alone. To explore the impact of additional spectral information, we adapted SOTA models, initially trained on 3 channels, to accept inputs from 7 channels. This adaptation involved extending the weights of the existing RGB channels to accommodate the newly introduced channels. The introduction of these additional channels appeared to enhance model performance significantly. However, it was noted that thermal channels did not contribute substantially to the improvement in this context. Our results indicate that incorporating a variety of bands can potentially enhance a model's performance in identifying irrigation patterns (see table \ref{table:channel}).

\begin{figure}
    \centering
    \includegraphics{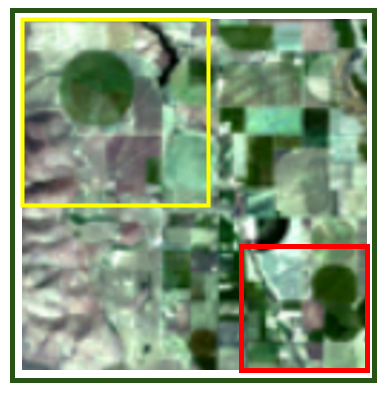}
    \caption{The image illustrates the variance in information levels across patches, highlighting the diverse data captured in different segments of the agricultural landscape. (Red Box: 64x64, Yellow: 128x128 Green: 256x256)}
    \label{fig:patch}
\end{figure}

\textbf{Progressive Training with Varying Patch Sizes.} In our methodology, we propose an initial training phase using smaller image patches that densely represent the class pixels associated with both sprinkler and flood irrigation types (see fig. \ref{fig:patch}). This focused approach allows the model to prioritize critical data while mastering the characteristics of different irrigation patterns. Training commenced with 64x64 patches, detailed in section \ref{sec:training}. We then implemented transfer learning to refine the model further on 128x128 patches. This step, as highlighted in Table \ref{table:varyingPatch}, offers substantial improvements over exclusively training with 128x128 patches. Continuing with this methodology, we extended training to 256x256 patches, achieving notable advancements in all evaluated metrics and surpassing competing models in efficiency, requiring fewer training epochs. Our strategy culminated in a Mean Intersection over Union (mIoU) score of 0.613 and an F1 score of 0.819. Through strategic application of transfer learning, utilizing the model's pretrained weights and incrementally introducing more detailed information, we significantly bolster the model's learning capacity. This progressive enlargement enables the model to initially recognize broad patterns before refining its understanding with more granular details as patch sizes increase (see fig. \ref{fig:predOutput}).

\begin{table}[htbp]
\centering
\caption{ Performance Comparison of State-of-the-Art Models Trained on 256x256 Patches vs Our Proposed Method}
\begin{tabular}{lccccc}
\toprule
Model & mIoU & Precision & Recall & F1 \\
\midrule
ResNet50\cite{he2015deep} & 0.398 & 0.631 & 0.639 & 0.635 \\
DeepLabv3+\cite{deeplab} & 0.321 & 0.595 & 0.585  & 0.590 \\
SegFormer\cite{SegFormer} & 0.323 & 0.613 & 0.612 & 0.612 \\
Inceptionv3\cite{inception} & 0.298 & 0.513 & 0.512 & 0.512 \\
Pre\-Landsat\cite{Stewart_TorchGeo_Deep_Learning_2022} & 0.523 & 0.681 & 0.679 & 0.680 \\
\midrule
\textbf{Ours} & \textbf{0.579} & \textbf{0.8273} & \textbf{0.8125} & \textbf{0.819} \\
\bottomrule
\end{tabular}
\label{table:SOTA}
\end{table}

\begin{table}[htbp]
\centering
\caption{Performance of ResNet50 varying different channels.}
\begin{tabular}{lccccc}
\toprule
Model & mIoU & F1\\
\midrule
RGB & 0.311 & 0.5991 \\
RGBN & 0.323 & 0.6509 \\
RGBNS1  & 0.358 & 0.6636 \\
RGBNS1S2  & 0.354 & 0.6529 \\
RGBNS1S2Th  & 0.322 & 0.6444  \\
\bottomrule
\end{tabular}
\label{table:channel}
\end{table}

\begin{table}[htbp]
\centering
\caption{Performance of ResNet50 varying different patch size with 6 channel(R:Red,G:Green,B:Blue,N:NIR,S:SWI1,S:SWI2)}
\begin{tabular}{@{}lcccc@{}}
\toprule
\multicolumn{1}{c}{Patch Size} & \multicolumn{2}{c}{Training Individually} & \multicolumn{2}{c}{Ours} \\ 
\cmidrule(lr){2-3} \cmidrule(lr){4-5}
& mIoU & F1  & mIoU & F1 \\ 
\midrule
64x64 & 0.295 & 0.587 & 0.295 & 0.587 \\
128x128 & 0.311 & 0.595 & \textbf{0.452} & \textbf{0.692} \\
256x256 & 0.401 & 0.579 & \textbf{0.613} & \textbf{0.819} \\
\bottomrule
\end{tabular}
\label{table:varyingPatch}
\end{table}

% \subsubsection{Quantitative Evaluation}
% \subsubsection{Qualitative Evaluation}

\section{Discussion}
This paper aims to tackle a critical issue in agriculture by developing methods to enhance water resource management and improve water quality in irrigated areas. To this end, we curated a dataset from Landsat remote sensing imagery, which offers an extensive historical record with 30m resolution, and employed the WRLU dataset for data labeling. While higher resolution data could potentially yield more precise results, Landsat's historical data enables an analysis of changes in irrigation practices over time, offering valuable insights for long-term agricultural planning and environmental sustainability. Our experimental findings demonstrate that along with incorporating multiple bands, progressively training the model with incremental information significantly improves performance.  Model tends to learn finer details in this approach. This approach is versatile and can be applied to various tasks involving Landsat imagery. Additionally, we investigated the influence of different spectral bands to provide insights on optimizing model training with limited data and computational resources. While our study primarily utilized patch sizes of 64x64, 128x128, and 256x256, exploring other dimensions and sequences could also prove beneficial. Our models serve as a foundation for generating additional data through iterative processes, including the creation of pseudo-labels and expert validation, which can substantially reduce the manual labor required for compiling extensive datasets. We compared the outputs of various SOTA models and found that models pre-trained on Landsat data offer a solid starting point for training on Landsat imagery. We hope our findings will serve as a foundational baseline for further research in irrigation mapping.

\section{Conclusion}
This paper proposes a transformative method of employing progressively increasing patch sizes during model training to address the challenges of irrigation mapping. By beginning with small patches to focus on detailed features and expanding to larger patches to capture broader landscape aspects, this approach effectively navigates the complex variabilities of agricultural fields equipped with diverse irrigation systems. Our results reveal a significant performance improvement, with a 20\% enhancement over state-of-the-art models, underscoring the efficacy of this strategy in refining model accuracy and robustness. Additionally, the integration of multiple spectral bands into the model training process has been analyzed, confirming their critical role in enhancing the model's capability to discern finer details crucial for accurate and effective mapping. This study also rigorously evaluates current state-of-the-art models. Overall, our findings not only demonstrate the potential of advanced machine learning techniques in improving irrigation mapping but also establish a new standard for the utilization of remote sensing data. This work paves the way for further research and development in the field, promising significant advancements in agricultural vision using remote sensing.

\section*{Acknowledgements}

This material is based upon work supported by the AI Research Institutes program supported by NSF and USDA-NIFA under the AI Institute: Agricultural AI for Transforming Workforce and Decision Support (AgAID) award No.~2021-67021-35344.

{
    \small
    \bibliographystyle{ieeenat_fullname}
    \bibliography{irrigation_Type_final}

\begin{thebibliography}{54}
\providecommand{\natexlab}[1]{#1}
\providecommand{\url}[1]{\texttt{#1}}
\expandafter\ifx\csname urlstyle\endcsname\relax
  \providecommand{\doi}[1]{doi: #1}\else
  \providecommand{\doi}{doi: \begingroup \urlstyle{rm}\Url}\fi

\bibitem[Abioye et~al.(2020)Abioye, Abidin, Mahmud, Buyamin, Ishak, Rahman, Otuoze, Onotu, and Ramli]{ABIOYE2020105441}
Emmanuel~Abiodun Abioye, Mohammad Shukri~Zainal Abidin, Mohd Saiful~Azimi Mahmud, Salinda Buyamin, Mohamad Hafis~Izran Ishak, Muhammad Khairie Idham~Abd Rahman, Abdulrahaman~Okino Otuoze, Patrick Onotu, and Muhammad Shahrul~Azwan Ramli.
\newblock A review on monitoring and advanced control strategies for precision irrigation.
\newblock \emph{Computers and Electronics in Agriculture}, 173:\penalty0 105441, 2020.

\bibitem[Agency(2020)]{epaNov2023}
US~Environmenta~Protection Agency.
\newblock Climate change impacts on agriculture and food supply.
\newblock \url{https://www.epa.gov/climateimpacts/climate-change-impacts-agriculture-and-food-supply}, 2020.
\newblock Accessed: 2024-03-24.

\bibitem[Badrinarayanan et~al.(2017)Badrinarayanan, Kendall, and Cipolla]{badrinarayanan2017segnet}
Vijay Badrinarayanan, Alex Kendall, and Roberto Cipolla.
\newblock Segnet: A deep convolutional encoder-decoder architecture for image segmentation.
\newblock \emph{IEEE Transactions on Pattern Analysis and Machine Intelligence}, 2, 2017.

\bibitem[Bai et~al.(2021)Bai, Bai, Li, and Liu]{app11115069}
Hao Bai, Tingzhu Bai, Wei Li, and Xun Liu.
\newblock A building segmentation network based on improved spatial pyramid in remote sensing images.
\newblock \emph{Applied Sciences}, 11\penalty0 (11), 2021.

\bibitem[Bazzi et~al.(2019)Bazzi, Baghdadi, Ienco, El~Hajj, Zribi, Belhouchette, Escorihuela, and Demarez]{bazzi2019mapping}
H. Bazzi, N. Baghdadi, D. Ienco, M. El~Hajj, M. Zribi, H. Belhouchette, M.~J. Escorihuela, and V. Demarez.
\newblock Mapping irrigated areas using sentinel-1 time series in catalonia, spain.
\newblock \emph{Remote Sensing}, 11\penalty0 (15):\penalty0 25, 2019.

\bibitem[Beyer et~al.(2023)Beyer, Izmailov, Kolesnikov, Caron, Kornblith, Zhai, Minderer, Tschannen, Alabdulmohsin, and Pavetic]{10205121}
L. Beyer, P. Izmailov, A. Kolesnikov, M. Caron, S. Kornblith, X. Zhai, M. Minderer, M. Tschannen, I. Alabdulmohsin, and F. Pavetic.
\newblock Flexivit: One model for all patch sizes.
\newblock In \emph{2023 IEEE/CVF Conference on Computer Vision and Pattern Recognition (CVPR)}, pages 14496--14506, Los Alamitos, CA, USA, 2023. IEEE Computer Society.

\bibitem[Bjorck et~al.(2021)Bjorck, Rappazzo, Shi, Brown-Lima, Dean, Fuller, and Gomes]{Bjorck2021AcceleratingES}
Johan Bjorck, Brendan~H. Rappazzo, Qinru Shi, Carrie Brown-Lima, Jennifer Dean, Angela~K. Fuller, and Carla~P. Gomes.
\newblock Accelerating ecological sciences from above: Spatial contrastive learning for remote sensing.
\newblock In \emph{AAAI Conference on Artificial Intelligence}, 2021.

\bibitem[Boutsioukis and Arias-Moliz(2022)]{boutsioukis2022present}
C Boutsioukis and MT Arias-Moliz.
\newblock Present status and future directions - irrigants and irrigation methods.
\newblock \emph{International Endodontic Journal}, 55\penalty0 (Suppl 3):\penalty0 588--612, 2022.

\bibitem[Chen et~al.(2018)Chen, Zhu, Papandreou, Schroff, and Adam]{deeplab}
Liang{-}Chieh Chen, Yukun Zhu, George Papandreou, Florian Schroff, and Hartwig Adam.
\newblock Encoder-decoder with atrous separable convolution for semantic image segmentation.
\newblock \emph{CoRR}, abs/1802.02611, 2018.

\bibitem[de~Geus et~al.(2020)de~Geus, Meletis, and Dubbelman]{geus2020fast}
Thijs de Geus, Panagiotis Meletis, and Gijs Dubbelman.
\newblock Fast panoptic segmentation network.
\newblock \emph{IEEE Robotics and Automation Letters}, 1:\penalty0 1--2, 6, 2020.

\bibitem[Dosovitskiy et~al.(2021)Dosovitskiy, Beyer, Kolesnikov, Weissenborn, Zhai, Unterthiner, Dehghani, Minderer, Heigold, Gelly, Uszkoreit, and Houlsby]{dosovitskiy2021image}
Alexey Dosovitskiy, Lucas Beyer, Alexander Kolesnikov, Dirk Weissenborn, Xiaohua Zhai, Thomas Unterthiner, Mostafa Dehghani, Matthias Minderer, Georg Heigold, Sylvain Gelly, Jakob Uszkoreit, and Neil Houlsby.
\newblock An image is worth 16x16 words: Transformers for image recognition at scale, 2021.

\bibitem[Feng et~al.(2023)Feng, Song, Yang, Zhang, and Jiao]{10187150}
Zhixi Feng, Liangliang Song, Shuyuan Yang, Xinyu Zhang, and Licheng Jiao.
\newblock Cross-modal contrastive learning for remote sensing image classification.
\newblock \emph{IEEE Transactions on Geoscience and Remote Sensing}, 61:\penalty0 1--13, 2023.

\bibitem[Fujinaga and Nakanishi(2023)]{10039322}
Takuya Fujinaga and Tsuneo Nakanishi.
\newblock Semantic segmentation of strawberry plants using deeplabv3+ for small agricultural robot.
\newblock In \emph{2023 IEEE/SICE International Symposium on System Integration (SII)}, pages 1--6, 2023.

\bibitem[He et~al.(2015)He, Zhang, Ren, and Sun]{he2015deep}
Kaiming He, Xiangyu Zhang, Shaoqing Ren, and Jian Sun.
\newblock Deep residual learning for image recognition, 2015.

\bibitem[Hoffer et~al.(2019)Hoffer, Weinstein, Hubara, Ben-Nun, Hoefler, and Soudry]{hoffer2019mix}
Elad Hoffer, Berry Weinstein, Itay Hubara, Tal Ben-Nun, Torsten Hoefler, and Daniel Soudry.
\newblock Mix \& match: training convnets with mixed image sizes for improved accuracy, speed and scale resiliency, 2019.

\bibitem[Howard(2018)]{Howard2018FastAI}
Jeremy Howard.
\newblock Fastai - progressive resizing, 2018.
\newblock Accessed: 03/24/2023.

\bibitem[Hu et~al.(2023)Hu, Huang, Ren, Zhang, Ji, and Cao]{Hu_2023_CVPR}
Jie Hu, Linyan Huang, Tianhe Ren, Shengchuan Zhang, Rongrong Ji, and Liujuan Cao.
\newblock You only segment once: Towards real-time panoptic segmentation.
\newblock In \emph{Proceedings of the IEEE/CVF Conference on Computer Vision and Pattern Recognition (CVPR)}, pages 17819--17829, 2023.

\bibitem[Karras et~al.(2018)Karras, Aila, Laine, and Lehtinen]{karras2018progressive}
Tero Karras, Timo Aila, Samuli Laine, and Jaakko Lehtinen.
\newblock Progressive growing of {GAN}s for improved quality, stability, and variation.
\newblock In \emph{International Conference on Learning Representations}, 2018.

\bibitem[Levidow et~al.(2014)Levidow, Zaccaria, Maia, Vivas, Todorovic, and Scardigno]{LEVIDOW201484}
Les Levidow, Daniele Zaccaria, Rodrigo Maia, Eduardo Vivas, Mladen Todorovic, and Alessandra Scardigno.
\newblock Improving water-efficient irrigation: Prospects and difficulties of innovative practices.
\newblock \emph{Agricultural Water Management}, 146:\penalty0 84--94, 2014.

\bibitem[Li et~al.(2023)Li, Wang, Zhao, Wang, and Zhong]{Li_Wang_Zhao_Wang_Zhong_2023}
Jingtao Li, Xinyu Wang, Hengwei Zhao, Shaoyu Wang, and Yanfei Zhong.
\newblock Anomaly segmentation for high-resolution remote sensing images based on pixel descriptors.
\newblock \emph{Proceedings of the AAAI Conference on Artificial Intelligence}, 37\penalty0 (4):\penalty0 4426--4434, 2023.

\bibitem[Lin et~al.(2017)Lin, Doll{\'a}r, Girshick, He, Hariharan, and Belongie]{fpn}
Tsung-Yi Lin, Piotr Doll{\'a}r, Ross Girshick, Kaiming He, Bharath Hariharan, and Serge Belongie.
\newblock Feature pyramid networks for object detection.
\newblock In \emph{Proceedings of the IEEE conference on computer vision and pattern recognition}, pages 2117--2125, 2017.

\bibitem[Liu et~al.(2023)Liu, Shi, Wang, and Zhong]{10378441}
Y. Liu, S. Shi, J. Wang, and Y. Zhong.
\newblock Seeing beyond the patch: Scale-adaptive semantic segmentation of high-resolution remote sensing imagery based on reinforcement learning.
\newblock In \emph{2023 IEEE/CVF International Conference on Computer Vision (ICCV)}, pages 16822--16832, Los Alamitos, CA, USA, 2023. IEEE Computer Society.

\bibitem[Marsocci et~al.(2023)Marsocci, Gonthier, Garioud, Scardapane, and Mallet]{Marsocci_2023_CVPR}
Valerio Marsocci, Nicolas Gonthier, Anatol Garioud, Simone Scardapane, and Cl\'ement Mallet.
\newblock Geomultitasknet: Remote sensing unsupervised domain adaptation using geographical coordinates.
\newblock In \emph{Proceedings of the IEEE/CVF Conference on Computer Vision and Pattern Recognition (CVPR) Workshops}, pages 2075--2085, 2023.

\bibitem[Mehta et~al.(2018)Mehta, Rastegari, Caspi, Shapiro, and Hajishirzi]{mehta2018espnet}
Sachin Mehta, Mohammad Rastegari, Anat Caspi, Linda Shapiro, and Hannaneh Hajishirzi.
\newblock Espnet: Efficient spatial pyramid of dilated convolutions for semantic segmentation.
\newblock In \emph{European Conference on Computer Vision}, 2018.

\bibitem[Mehta et~al.(2019)Mehta, Rastegari, Shapiro, and Hajishirzi]{mehta2019espnetv2}
Sachin Mehta, Mohammad Rastegari, Linda Shapiro, and Hannaneh Hajishirzi.
\newblock Espnetv2: A light-weight, power efficient, and general purpose convolutional neural network.
\newblock In \emph{Proceedings of the IEEE/CVF Conference on Computer Vision and Pattern Recognition}, 2019.

\bibitem[Osowski(2020)]{msu2020irrigation}
Val Osowski.
\newblock Long-term mapping key to effective management of irrigated areas.
\newblock \url{https://msutoday.msu.edu/news/2020/long-term-mapping-key-to-effective-management-of-irrigated-areas}, 2020.
\newblock Accessed: 2024-03-24.

\bibitem[Paszke et~al.(2016)Paszke, Chaurasia, Kim, and Culurciello]{paszke2016enet}
Adam Paszke, Abhishek Chaurasia, Sangpil Kim, and Eugenio Culurciello.
\newblock Enet: A deep neural network architecture for real-time semantic segmentation.
\newblock \emph{arXiv preprint arXiv:1606.02147}, 2016.

\bibitem[Quintana et~al.(2023)Quintana, Li, Vancamberg, Mougeot, Desolneux, and Muller]{bioengineering10050534}
Gonzalo~Iñaki Quintana, Zhijin Li, Laurence Vancamberg, Mathilde Mougeot, Agnès Desolneux, and Serge Muller.
\newblock Exploiting patch sizes and resolutions for multi-scale deep learning in mammogram image classification.
\newblock \emph{Bioengineering}, 10\penalty0 (5), 2023.

\bibitem[Raei et~al.(2022)Raei, {Akbari Asanjan}, Nikoo, Sadegh, Pourshahabi, and Adamowski]{RAEI2022106977}
Ehsan Raei, Ata {Akbari Asanjan}, Mohammad~Reza Nikoo, Mojtaba Sadegh, Shokoufeh Pourshahabi, and Jan~Franklin Adamowski.
\newblock A deep learning image segmentation model for agricultural irrigation system classification.
\newblock \emph{Computers and Electronics in Agriculture}, 198:\penalty0 106977, 2022.

\bibitem[Ronneberger et~al.(2015)Ronneberger, Fischer, and Brox]{unet}
Olaf Ronneberger, Philipp Fischer, and Thomas Brox.
\newblock U-net: Convolutional networks for biomedical image segmentation.
\newblock \emph{CoRR}, abs/1505.04597, 2015.

\bibitem[Salmon et~al.(2015)Salmon, Friedl, Frolking, Wisser, and Douglas]{salmon2015global}
J.~M. Salmon, M.~A. Friedl, S. Frolking, D. Wisser, and E.~M. Douglas.
\newblock Global rain-fed, irrigated, and paddy croplands: A new high resolution map derived from remote sensing, crop inventories and climate data.
\newblock \emph{International Journal of Applied Earth Observation and Geoinformation}, 38:\penalty0 321–334, 2015.

\bibitem[Saraiva et~al.(2020)Saraiva, Protas, Salgado, and Souza]{saraiva2020automatic}
M. Saraiva, E. Protas, M. Salgado, and C. Souza.
\newblock Automatic mapping of center pivot irrigation systems from satellite images using deep learning.
\newblock \emph{Remote Sensing}, 12\penalty0 (3):\penalty0 14, 2020.

\bibitem[Scheibenreif et~al.(2022)Scheibenreif, Hanna, Mommert, and Borth]{9857009}
Linus Scheibenreif, Joëlle Hanna, Michael Mommert, and Damian Borth.
\newblock Self-supervised vision transformers for land-cover segmentation and classification.
\newblock In \emph{2022 IEEE/CVF Conference on Computer Vision and Pattern Recognition Workshops (CVPRW)}, pages 1421--1430, 2022.

\bibitem[Shen et~al.(2023)Shen, Lin, Xu, and Wu]{Shen2023}
X. Shen, L. Lin, X. Xu, and S. Wu.
\newblock Effects of patchwise sampling strategy to three-dimensional convolutional neural network-based alzheimer's disease classification.
\newblock \emph{Brain Sci}, 13\penalty0 (2):\penalty0 254, 2023.

\bibitem[Shorten and Khoshgoftaar(2019)]{Shorten2019}
Connor Shorten and Taghi~M. Khoshgoftaar.
\newblock A survey on image data augmentation for deep learning.
\newblock \emph{Journal of Big Data}, 6\penalty0 (1):\penalty0 60, 2019.

\bibitem[Siebert et~al.(2015)Siebert, Kummu, Porkka, Doll, Ramankutty, and Scanlon]{siebert2015globaldata}
S. Siebert, M. Kummu, M. Porkka, P. Doll, N. Ramankutty, and B.~R. Scanlon.
\newblock A global data set of the extent of irrigated land from 1900 to 2005.
\newblock \emph{Hydrology and Earth System Sciences}, 19\penalty0 (3):\penalty0 1521–1545, 2015.

\bibitem[Sikdar et~al.(2023)Sikdar, Udupa, Gurunath, and Sundaram]{10208987}
Aniruddh Sikdar, Sumanth Udupa, Prajwal Gurunath, and Suresh Sundaram.
\newblock Deepmao: Deep multi-scale aware overcomplete network for building segmentation in satellite imagery.
\newblock In \emph{2023 IEEE/CVF Conference on Computer Vision and Pattern Recognition Workshops (CVPRW)}, pages 487--496, 2023.

\bibitem[Stewart et~al.(2022)Stewart, Robinson, Corley, Ortiz, Lavista~Ferres, and Banerjee]{Stewart_TorchGeo_Deep_Learning_2022}
Adam~J. Stewart, Caleb Robinson, Isaac~A. Corley, Anthony Ortiz, Juan~M. Lavista~Ferres, and Arindam Banerjee.
\newblock {TorchGeo}: Deep learning with geospatial data.
\newblock In \emph{Proceedings of the 30th International Conference on Advances in Geographic Information Systems}, pages 1--12, Seattle, Washington, 2022. Association for Computing Machinery.

\bibitem[Szegedy et~al.(2015)Szegedy, Vanhoucke, Ioffe, Shlens, and Wojna]{inception}
Christian Szegedy, Vincent Vanhoucke, Sergey Ioffe, Jonathon Shlens, and Zbigniew Wojna.
\newblock Rethinking the inception architecture for computer vision.
\newblock \emph{CoRR}, abs/1512.00567, 2015.

\bibitem[Tang et~al.(2021)Tang, Arvor, Corpetti, and Tang]{tang2021mapping}
J.~W. Tang, D. Arvor, T. Corpetti, and P. Tang.
\newblock Mapping center pivot irrigation systems in the southern amazon from sentinel-2 images.
\newblock \emph{Water}, 13\penalty0 (3):\penalty0 298, 2021.

\bibitem[Todescato et~al.(2024)Todescato, Garcia, Balreira, and Carbonera]{TODESCATO2024121116}
Matheus~V. Todescato, Luan~F. Garcia, Dennis~G. Balreira, and Joel~L. Carbonera.
\newblock Multiscale patch-based feature graphs for image classification.
\newblock \emph{Expert Systems with Applications}, 235:\penalty0 121116, 2024.

\bibitem[Touvron et~al.(2019{\natexlab{a}})Touvron, Vedaldi, Douze, and Jegou]{NEURIPS2019_d03a857a}
Hugo Touvron, Andrea Vedaldi, Matthijs Douze, and Herve Jegou.
\newblock Fixing the train-test resolution discrepancy.
\newblock In \emph{Advances in Neural Information Processing Systems}. Curran Associates, Inc., 2019{\natexlab{a}}.

\bibitem[Touvron et~al.(2019{\natexlab{b}})Touvron, Vedaldi, Douze, and J{\'e}gou]{Touvron2019TrainTestResolution}
Hugo Touvron, Andrea Vedaldi, Matthijs Douze, and Herv{\'e} J{\'e}gou.
\newblock Fixing the train-test resolution discrepancy.
\newblock \emph{CoRR}, abs/1906.06423, 2019{\natexlab{b}}.

\bibitem[{U.S. Geological Survey}(2023)]{usgs_landsat}
{U.S. Geological Survey}.
\newblock {What is the Landsat Satellite Program and Why is it Important?}
\newblock \url{https://www.usgs.gov/faqs/what-landsat-satellite-program-and-why-it-important}, 2023.
\newblock Accessed in 2024-03-24.

\bibitem[{van Noord} and Postma(2017)]{VANNOORD2017583}
Nanne {van Noord} and Eric Postma.
\newblock Learning scale-variant and scale-invariant features for deep image classification.
\newblock \emph{Pattern Recognition}, 61:\penalty0 583--592, 2017.

\bibitem[Wang et~al.(2020)Wang, Sun, Cheng, Jiang, Deng, Zhao, Liu, Mu, Tan, Wang, Liu, and Xiao]{wang2020deep}
Jingdong Wang, Ke Sun, Tianheng Cheng, Borui Jiang, Chaorui Deng, Yang Zhao, Dong Liu, Yadong Mu, Mingkui Tan, Xinggang Wang, Wenyu Liu, and Bin Xiao.
\newblock Deep high-resolution representation learning for visual recognition, 2020.

\bibitem[Xie et~al.(2021{\natexlab{a}})Xie, Wang, Yu, Anandkumar, Alvarez, and Luo]{SegFormer}
Enze Xie, Wenhai Wang, Zhiding Yu, Anima Anandkumar, Jose~M Alvarez, and Ping Luo.
\newblock Segformer: Simple and efficient design for semantic segmentation with transformers.
\newblock In \emph{Neural Information Processing Systems (NeurIPS)}, 2021{\natexlab{a}}.

\bibitem[Xie et~al.(2021{\natexlab{b}})Xie, Wang, Yu, Anandkumar, Alvarez, and Luo]{xie2021segformer}
Enze Xie, Wenhai Wang, Zhiding Yu, Anima Anandkumar, Jose~M Alvarez, and Ping Luo.
\newblock Segformer: Simple and efficient design for semantic segmentation with transformers.
\newblock In \emph{Advances in Neural Information Processing Systems}, 2021{\natexlab{b}}.

\bibitem[Yadav et~al.(2014)Yadav, Maheshwari, and Agarwal]{6968381}
Garima Yadav, Saurabh Maheshwari, and Anjali Agarwal.
\newblock Contrast limited adaptive histogram equalization based enhancement for real time video system.
\newblock In \emph{2014 International Conference on Advances in Computing, Communications and Informatics (ICACCI)}, pages 2392--2397, 2014.

\bibitem[Yu et~al.(2021)Yu, Gao, Wang, Yu, Shen, and Sang]{yu2021bisenetv2}
Changqian Yu, Changxin Gao, Jingbo Wang, Gang Yu, Chunhua Shen, and Nong Sang.
\newblock Bisenetv2: Bilateral network with guided aggregation for real-time semantic segmentation.
\newblock \emph{International Journal of Computer Vision}, 2, 2021.

\bibitem[Zhang et~al.(2023)Zhang, Jiang, Zheng, and Yao]{Zhang2023}
Hua Zhang, Zhengang Jiang, Guoxun Zheng, and Xuekun Yao.
\newblock Semantic segmentation of high-resolution remote sensing images with improved u-net based on transfer learning.
\newblock \emph{International Journal of Computational Intelligence Systems}, 16\penalty0 (1):\penalty0 181, 2023.

\bibitem[Zhang and Zhang(2023)]{ZHANG2023107511}
Shanwen Zhang and Chuanlei Zhang.
\newblock Modified u-net for plant diseased leaf image segmentation.
\newblock \emph{Computers and Electronics in Agriculture}, 204:\penalty0 107511, 2023.

\bibitem[Zhao et~al.(2018)Zhao, Qi, Shen, Shi, and Jia]{zhao2018icnet}
Hengshuang Zhao, Xiaojuan Qi, Xiaoyong Shen, Jianping Shi, and Jiaya Jia.
\newblock Icnet for real-time semantic segmentation on high-resolution images.
\newblock In \emph{European Conference on Computer Vision}, 2018.

\bibitem[Zhu et~al.(2023)Zhu, Jiang, Dong, Wu, and Ma]{10146268}
Zhijia Zhu, Mingkun Jiang, Jun Dong, Shuang Wu, and Fan Ma.
\newblock Pd-segnet: Semantic segmentation of small agricultural targets in complex environments.
\newblock \emph{IEEE Access}, 11:\penalty0 90214--90226, 2023.

\end{thebibliography}
}


\end{document}